\documentclass[conference，anonymous]{IEEEtran}
\IEEEoverridecommandlockouts
\usepackage{graphicx} 
\usepackage{caption}
\usepackage{amsmath} 

\usepackage{ulem}
\usepackage{xcolor}
\usepackage{multirow}

\DeclareMathAlphabet\mathbfcal{OMS}{cmsy}{b}{n}

\title{Key Information Retrieval to Classify the Unstructured Data Content of  Preferential Trade Agreements}
\author{Jiahui Zhao$^{1}$, Ziyi Meng$^{1}$, Stepan Gordeev$^{1}$, Zijie Pan$^{1}$, Dongjin Song$^{1}$, Sandro Steinbach$^{2}$, Caiwen Ding$^{1}$ \\
$^{1}$University of Connecticut, $^{2}$North Dakota State University\\
{\tt\small \{jiahui.zhao, nicole.meng, stepan.gordeev, zijie.pan, dongjin.song, caiwen.ding\}@uconn.edu, sandro.steinbach@ndsu.edu}
\vspace{-8mm}}
\date{August 2023}

\begin{document}

\maketitle
\begin{abstract}
With the rapid proliferation of textual data, predicting long texts has emerged as a significant challenge in the domain of natural language processing. Traditional text prediction methods encounter substantial difficulties when grappling with long texts, primarily due to the presence of redundant and irrelevant information, which impedes the model's capacity to capture pivotal insights from the text. To address this issue, we introduce a novel approach to long-text classification and prediction. Initially, we employ embedding techniques to condense the long texts, aiming to diminish the redundancy therein. Subsequently,the Bidirectional Encoder Representations from Transformers (BERT) embedding method is utilized for text classification training. Experimental outcomes indicate that our method realizes considerable performance enhancements in classifying long texts of Preferential Trade Agreements. Furthermore, the condensation of text through embedding methods not only augments prediction accuracy but also substantially reduces computational complexity. Overall, this paper presents a strategy for long-text prediction, offering a valuable reference for researchers and engineers in the natural language processing sphere.

\end{abstract}

\maketitle

\section{\textbf{Introduction}}
\vspace{-2pt}

The rapid progress in information technology and the widespread use of extensive datasets pose a significant challenge for large language models. Handling voluminous documents becomes increasingly intricate, particularly when there is a need to extract specific information from these extensive textual sources.

The application we focus on is classifying whether the text of a given preferential trade agreement (PTA) contains a particular provision. Modern PTAs are increasingly complex and diverse: systematic classification of provisions included in PTAs is crucial for the field of international trade governance to conduct quantitative analysis of their determinants and impact \cite{limaoPreferentialTradeAgreements2016}. Design of International Trade Agreements (DESTA) and Deep Trade Agreements (DTA) are two datasets that manually classify the presence of several hundred provisions in a subset of existing PTAs \cite{durDesignInternationalTrade2014, mattooEvolutionDeepTrade2020}. However, the manual approach to classification adopted by these two datasets limits the number of PTAs and provisions they can feasibly classify, limiting the breadth and depth of any empirical analysis using them. Therefore, we aim to automate the classification task with NLP, extending the classification of all provisions defined by DTA and DESTA to all PTAs listed in the NSF-Kellogg Institute Data Base on Economic Integration Agreements, the most complete database of texts of trade agreements \cite{nsf-kellogginstituteDatabaseEconomicIntegration2021}. Technically, our objective is to classify whether a long text document (each PTA) contains an answer to each ``question'' (the presence of a given provision). If the provision is present, the label is set to 1; if not, the label is set to 0. Our task is predicting this ``label''. 

When dealing with the texts of Preferential Trade Agremeents, which may extend beyond 100,000 words, we are confronted with complex processing issues. To efficiently mitigate this complexity, we employ embedding methods~\cite{zhao2020connecting, zhao2020knowledge, jiang2019deltadou} to select the top-k relevant key paragraphs from each article, effectively shortening the length of the documents. To ensure that meaningful paragraphs are not erroneously segmented, we adopt a sliding window method for segmenting the documents.

Moreover, we employ a method to extract keywords from the questions to obtain the ground truth, i.e., the answers judged as correct through manual assessment, and then compare these results with the top-k paragraphs to confirm that the selected paragraphs genuinely encompass the answers to the questions.

After these treatments, these paragraphs are fed as input data into an NLP model to perform the classification task, thus achieving the prediction of whether the document contains the answer to the question or not.

In the following sections, we will elaborate on our methodology, experimental setup, the results obtained, and comparisons with other methods.

\begin{figure*}
    \centering
    \includegraphics[width=1\textwidth]{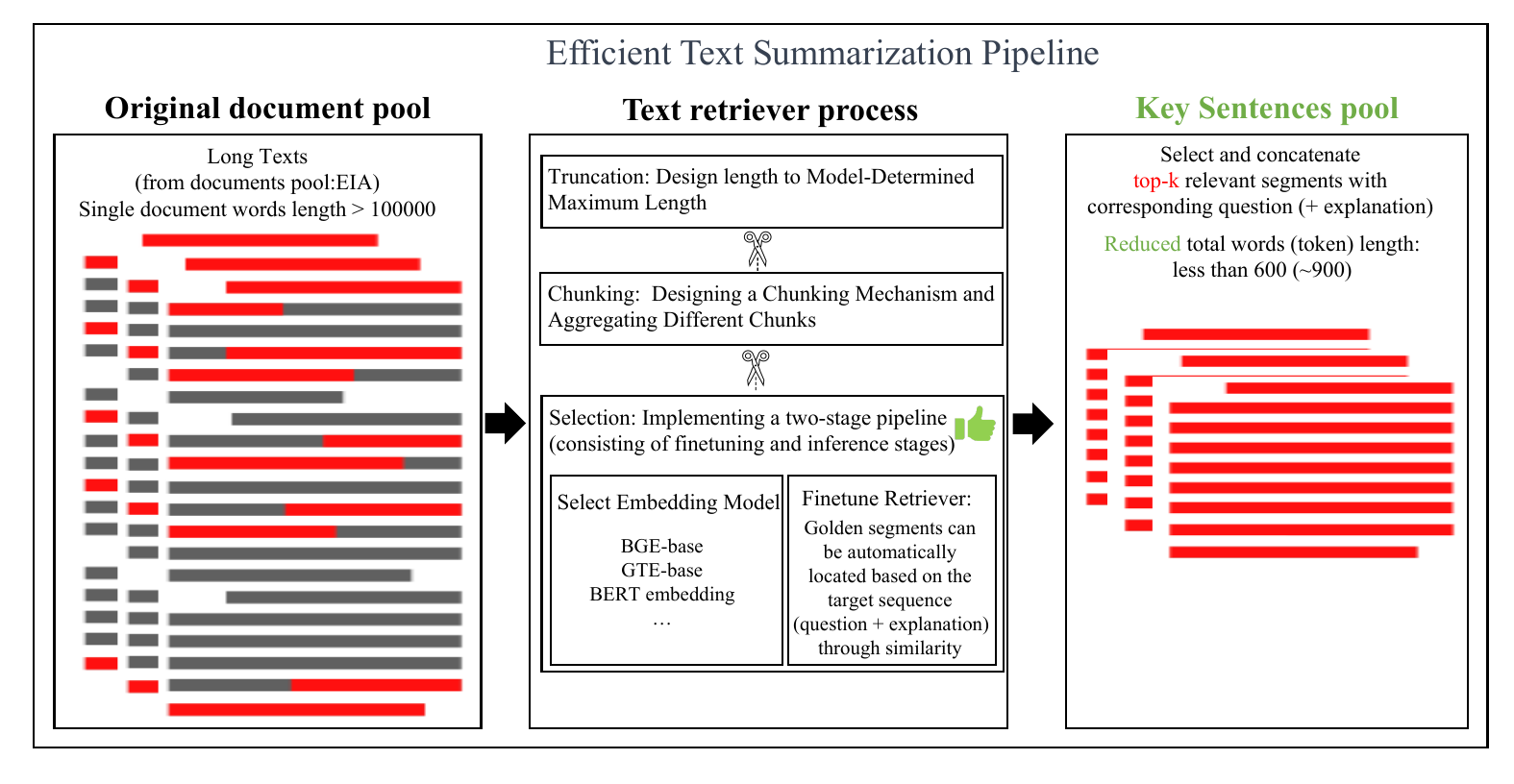}
    \vspace{-8mm}
    \caption{Overview of the Long Document Classification Pipeline. The process begins with an original document pool combined with DESTADATA to initiate the text retriever process, which involves truncation, chunking, and selection stages, using models like BGE-base, GTE-base, and BERT embeddings. The refined data then passes through a Ground Truth Finding Algorithm and is further reduced to under 600 tokens. This optimized content, alongside question explanations, feeds into a fine-tuning stage on a classification task, outputting a label that determines if the long document contains the answer to the input question.}
    \vspace{-4mm}
    \label{fig:ot_1}
\end{figure*}

\section{\textbf{Methodology}}

\subsection{Document Segmentation and Key Paragraph Extraction}

The extraction of critical information from extensive documents, especially when they surpass 100,000 words, necessitates a meticulous segmentation process to safeguard the contextual and semantic integrity of the text.

\subsubsection{Context-Aware Text Partitioning Method}

To more effectively handle documents that may exceed 100,000 words, we have adopted a new document segmentation method called "Context-Aware Text Partitioning" (CATP). Unlike the traditional sliding window method, which simply slides a fixed-size window over the document to extract text segments, CATP intelligently determines the size of the segments, ensuring the semantic integrity and contextual coherence of the text.

Specifically, this method can automatically adjust the window size based on the structure and content of the text, such as continuing until the end of a sentence or paragraph, to ensure the completeness and coherence of information. Additionally, CATP introduces the concept of overlapping windows, allowing for a certain proportion of overlap between consecutive windows, thus ensuring that no contextual information is lost during segmentation. This method allows us to effectively shorten document lengths without sacrificing text quality and improve accuracy and computational efficiency in subsequent natural language processing models for classification and prediction~\cite{accelgcn2023}.

To elaborate on the CATP method, we define a context-based segmentation function \( f \). This function takes the position \( p \) in the document \( D \) as input and determines the window size \( w \) for segmentation:

\begin{equation}
w = f(p, D)
\end{equation}

Here, \( p \) is the current position, \( D \) is the entire document, and \( w \) is the window size starting from position \( p \). We define \( f \) to ensure that each window preserves semantically complete information as much as possible. For instance, if \( p \) points to the beginning of a paragraph, \( w \) could extend to the end of that paragraph.

Furthermore, we define an overlap function \( g \), which decides the degree of overlap \( o \) between adjacent windows based on contextual necessity:

\begin{equation}
o = g(w, D)
\end{equation}

In this formula, \( o \) is the overlap size determined based on the size of the previous window \( w \) and the entire document \( D \). In this way, we ensure that continuous parts of the document are semantically coherent and reduce information loss due to window segmentation.

By combining these two functions, the CATP method adapts to different document structures and complexities of content, thereby optimizing the text segmentation process.

\begin{figure}
    \centering
    \includegraphics[width=1\linewidth]{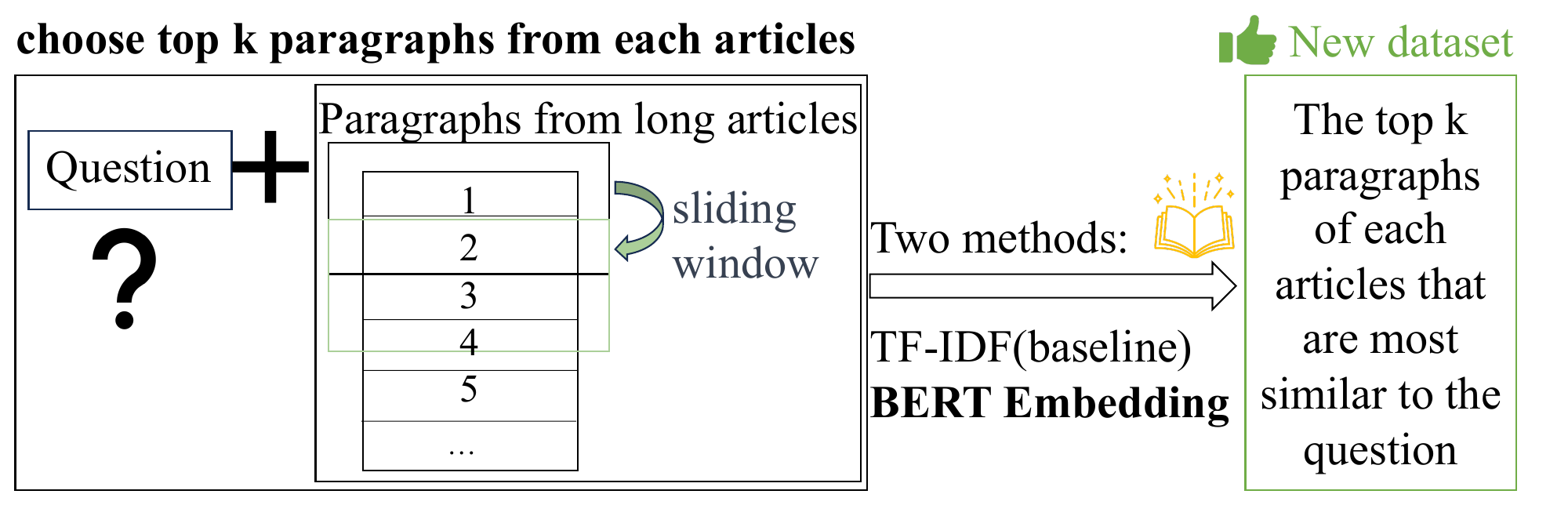}
    
  \captionof{figure}{Workflow of the Context-Aware Text Partitioning Method for dataset creation. The process compares a query question with long articles using two techniques: TF-IDF and BERT Embedding Methods. Each method selects the top k(10) relevant paragraphs, which are then assessed through a sliding window to establish ground truth data. The resulting dataset comprises paragraphs ranked by their similarity to the query.}
    \label{fig:ot_2}
    \vspace{-4pt}
\end{figure}

\subsubsection{Embedding Method and Key Paragraph Selection}

Initially, Term Frequency - Inverse Document Frequency(TF-IDF) was utilized as a baseline method, converting each extracted paragraph into vector form. Further, we found that utilizing BERT embeddings as the vector representation of paragraphs could achieve superior performance. Cosine similarity between the vector of the question and each paragraph vector is computed, selecting the top-k most relevant paragraphs for further analysis.

\[
\text{Similarity}(q, p) = \frac{q \cdot p}{\|q\| \|p\|}
\]

\subsection{Keyword Extraction and Ground Truth Acquisition}

Using a keyword extraction method based on TF-IDF, we determine the key information within the question and through manual assessment compare it with the pre-selected paragraphs, ensuring that the chosen paragraphs indeed contain the answer to the question.

\begin{figure*}
    \centering
    \includegraphics[width=1\textwidth]{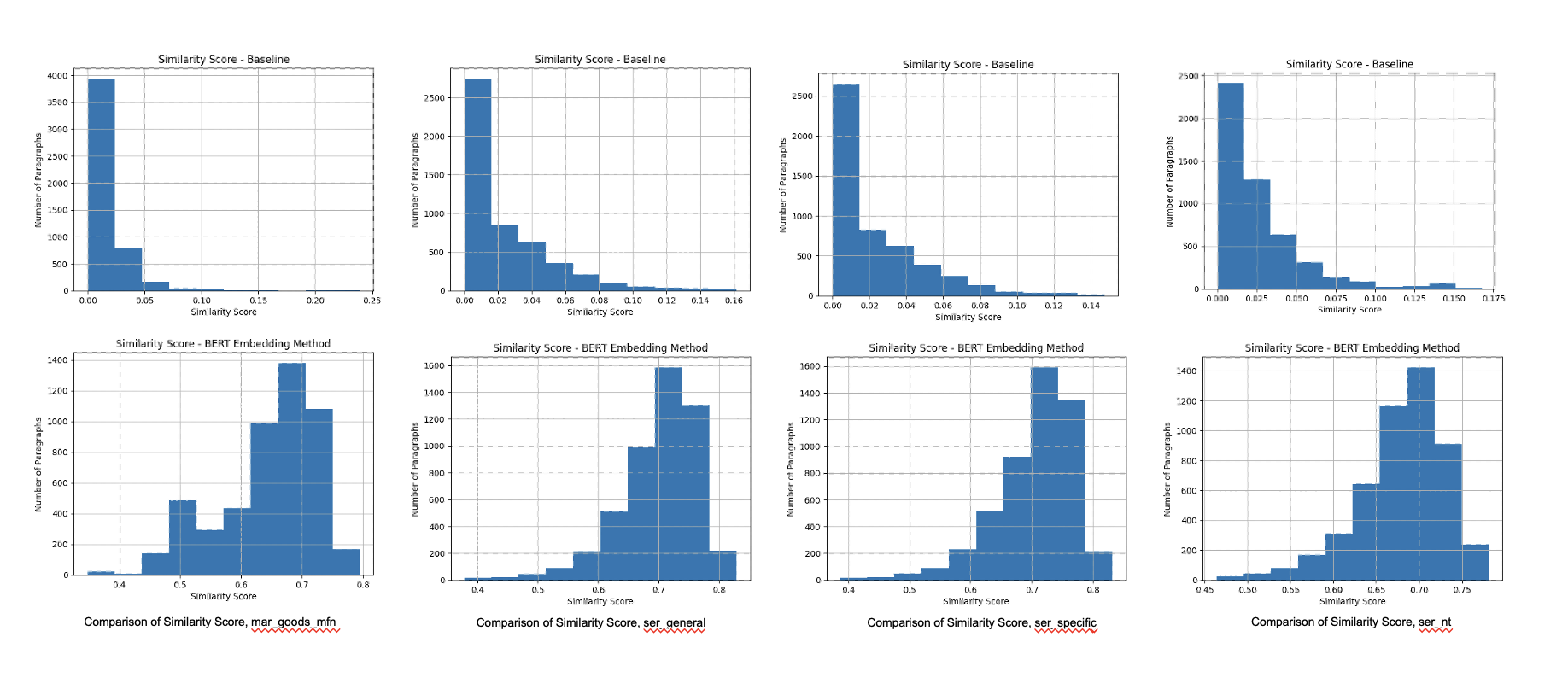}
    \vspace{-8mm}
    \caption{Sampled similiarity score improvement of BERT embedding method in comparison with the baseline(TF-IDF) model}
    \vspace{-4mm}
    \label{fig:ot_3}
\end{figure*}

\section{\textbf{Experimental Setup}}

\subsection{Embedding Method Experiments}

In order to address the challenges posed by lengthy documents in natural language processing (NLP) tasks, such as text classification and information retrieval, it becomes imperative to strike a balance between accuracy and model efficiency. Many real-world documents, especially those in academic research, legal, or technical fields, can be quite extensive, making them less than ideal as direct inputs for NLP models.

One approach to dealing with this issue is to employ techniques like TF-IDF (Term Frequency-Inverse Document Frequency) and BERT (Bidirectional Encoder Representations from Transformers) Embedding Method. These methods are introduced with the primary goal of identifying and extracting the most pertinent and contextually relevant portions of a document, which can then be used as input for subsequent analysis or classification tasks.

\begin{itemize}
\item \textbf{TF-IDF}: TF-IDF is a widely-used technique that evaluates the importance of words in a document within the context of a larger corpus. By assigning weights to words based on their frequency in a document and their rarity in the overall corpus, TF-IDF can help identify key terms and phrases that are indicative of the document's content. This approach is effective for tasks like document retrieval, keyword extraction, and text summarization. It allows for the selection of the most informative paragraphs or sections of a document.~\cite{ramos2003using}

\item \textbf{BERT}: a state-of-the-art transformer-based model, has revolutionized the field of NLP by providing contextually rich word embeddings. BERT's ability to capture the meaning of words in context has made it a valuable tool for understanding the content of lengthy documents. In the context of document analysis, BERT can be used to embed entire documents or specific paragraphs. Once the document is embedded, it becomes easier to compare and assess its relevance to specific topics or categories. By leveraging BERT, we can identify the most contextually significant parts of a document and use them as input for downstream tasks, such as document classification or sentiment analysis.
\end{itemize}

By integrating TF-IDF and BERT-based techniques into the workflow, we can efficiently filter out and retain only the most informative segments of extensive documents and use TF-IDF as a baseline model in order to assess the performance of BERT~\cite{devlin2018bert}.

\subsubsection{Experimental Dataset}
To validate the effectiveness of our embedding methods in selecting the most crucial paragraphs that contain answers to specific questions, a series of individual experiments were conducted. 

First, a set of questions was randomly selected and for each of these questions, we went through a meticulous process to create a dedicated experimental dataset. This dataset includes key components such as the file name, the question itself, and the paragraphs that potentially contain answers to the question. The process of selecting these relevant paragraphs involved a combination of algorithmic and manual verification, ensuring the highest level of precision.

By generating these datasets, we aimed to establish a robust framework for evaluating the performance of our embedding methods. A critical aspect of this evaluation is to measure the extent to which our methods successfully identify the most pertinent information. To accomplish this, we compared the output of our methods, which ranked the paragraphs, with a set of manually checked paragraphs containing the correct answers.

The primary assessment criterion was to determine whether the manually checked paragraphs were consistently ranked within the top-k returned paragraphs produced by our embedding methods. The "top-k" refers to the k highest-ranked paragraphs according to our algorithm. By employing this comparative analysis, we can quantify the effectiveness and reliability of our embedding techniques in isolating and prioritizing the paragraphs that contain the sought-after answers.

This rigorous experimental dataset creation process, combining automated algorithms and manual verification, ensures that the quality and accuracy of our results are rigorously assessed.

\subsubsection{Experimental setup} 
To establish a baseline for comparing the performance of our embedding methods, we opted for a widely recognized mathematical model known as TF-IDF (Term Frequency-Inverse Document Frequency). In this context, we leveraged the Natural Language Toolkit (NLTK) to implement the TF-IDF model. The purpose of incorporating this baseline model is to provide a reference point against which we can evaluate the effectiveness and improvements introduced by our BERT embedding techniques.

One distinctive feature of the TF-IDF model is that it doesn't rely on complex vectorization optimization methods. Instead, it calculates the importance of words within documents based on term frequency and inverse document frequency. By utilizing TF-IDF through NLTK, we were able to generate a set of top-k paragraphs relevant to specific questions, serving as our baseline for accuracy assessment.

In our evaluation process, one key metric to assess the performance is the similarity score. As previously defined in Section 2 of this document, the similarity score measures the cosine similarity between the question and the selected paragraph. This metric quantifies the extent of relevance between the query and the chosen paragraph, providing a basis for assessing the quality of our results.

\subsubsection{Maximizing Similarity Score}
In our quest to optimize document analysis, we focus on achieving high similarity scores as a key metric. We conducted experiments comparing the baseline TF-IDF model with the advanced BERT embedding method on randomly selected questions. In Figure 3, the results reveal a substantial increase in similarity scores with BERT, reinforcing its preference as our document analysis model.

\subsection{Optimize the value of k}
To enhance our information retrieval system's adaptability and effectiveness, we face the challenge of dynamically optimizing the parameter "k" for selecting top-ranked paragraphs without prior knowledge of the text file content or specific questions. "k" signifies the number of paragraphs chosen for analysis, tailored to each question's performance measured by Similarity Score to maximize precision and relevance.

In our experiments, we systematically explored various "k" values, such as 5, 10, and 15 for each question, aiming to identify the optimal "k" value that produces the highest similarity scores. This metric, crucial for precision, gauges the correct identification proportion among the top-ranked paragraphs. Our engineering approach involves evaluating each "k" value's performance for every question, enabling us to intelligently adapt our system to nuances, ultimately reducing the paragraphs needed for analysis.

This dynamic optimization approach allows fine-tuning of our information retrieval process for peak performance. Consistently selecting the "k" value associated with the best results enhances the efficiency and effectiveness of our document analysis, streamlining the pipeline and demonstrating superior results in comprehensive experiments.

\section{\textbf{Experiment Results}}

To assess our BERT embedding method's efficacy, we assessed the improved similarity scores for all selected questions. Each question demonstrated a significant enhancement in the top k paragraph similarity scores with the application of BERT, as depicted in Figure 3.

Upon careful examination of the results, it becomes evident that our BERT embedding method excelled in accurately identifying the top k most relevant paragraphs, consistently achieving a substantial improvement of over 50\% for every question tested. This performance superiority was notably evident when compared to the baseline model. To ensure a controlled study environment, we consistently set k to be 5 across all 10 questions. This choice reinforces our confidence in the BERT method as a robust embedding technique for the purpose of selecting relevant paragraphs in the context of lengthy documents.

our experiment confirms the effectiveness of our BERT embedding method in efficiently identifying key information in lengthy documents. When faced with extensive texts and questions, our method adeptly shortens the document to a crucial portion, showcasing its potential utility for training NLP models to analyze lengthy documents more efficiently in the future.

\section{\textbf{Conclusion}}
In summary, our targeted pre-selection of key paragraphs using the BERT embedding method provides a significant advantage over conventional approaches. This not only enhances runtime but also improves overall accuracy by streamlining the training process, making it more effective and resource-efficient compared to indiscriminate consideration of the entire document. 
Potential future direction of proposed work can be further incorporating GPU acceleration design into the Transformer embedding system~\cite{xie2023accel, wang2023digital, peng2023maxk, sheng2023muffin}, using large language model as a auxiliary tool to enhance the retrieval performance~\cite{sun2023chatgpt, thorat2023advanced, wu2024switchtab}, and enabling privacy design in the retrieval system~\cite{peng2023rrnet, luo2023aq2pnn, peng2023autorep, wu2021federated, peng2023lingcn, wu2021federated2, peng2023pasnet}.

\section*{Acknowledgement}
This work was in part supported by the USDA-NIFA Agriculture and Food Research Initiative Program (Award No.: 2022-67023-36399).

\bibliographystyle{unsrt}
\bibliography{references}

\end{document}